%% file: paper.tex
\documentclass[]{TEAI}
\usepackage{helvet}

\usepackage{amsmath} 
\usepackage{natbib}
\usepackage{graphicx}
\usepackage{subcaption} 

\usepackage[toc,page,header]{appendix}
\usepackage[utf8]{inputenc} 
\usepackage[T1]{fontenc}    
\usepackage{hyperref}       
\usepackage{url}            
\usepackage{booktabs}       
\usepackage{lmodern}        
\usepackage{amsfonts}       
\usepackage{nicefrac}       
\usepackage{microtype}      
\usepackage{wrapfig}

\usepackage{amssymb}  
\usepackage{fontawesome}  
\usepackage{url}  

\usepackage{titletoc}

\usepackage{tikz}  
\usepackage{comment}  
\usepackage{tabularx}  
\usepackage{booktabs}  

\usepackage{minitoc}

\usepackage{booktabs}
\usepackage{array}
\usepackage{etoolbox}

\definecolor{lightblue}{RGB}{200, 230, 255}  
\definecolor{headerblue}{RGB}{150, 200, 255} 

\usepackage{pgfplots}
\usepackage[utf8]{inputenc} 
\usepackage[T1]{fontenc}   
\usepackage{hyperref}       
\usepackage{url}            
\usepackage{booktabs}       
\usepackage{amsfonts}       
\usepackage{nicefrac}       
\usepackage{microtype}      
\usepackage{xcolor}         
\usepackage{graphicx}
\usepackage{float}
\usepackage{comment}
\usepackage{multirow} 
\usepackage{amsmath} 
\usepackage{makecell} 
\usepackage{siunitx}  
\usepackage{tikz}
\usepackage{pgf-pie} 
\usepackage{subcaption}
\usepackage{wrapfig}
\usepackage[export]{adjustbox}

\usepackage{ragged2e}      
\usepackage{tabularx}       
\usepackage{array}          
\usepackage{caption}     
\usepackage{enumitem}
\usepackage{pifont}
\usepackage[hang,flushmargin]{footmisc} 

\usepackage{tcolorbox}

\tcbuselibrary{breakable}  
\tcbuselibrary{skins}      

\usepackage{tabularx}
\usepackage{listings}
\usepackage{wrapfig}

\usepackage{float}
\usepackage{xspace}
\usepackage{graphicx} 
\usepackage{tabularx} 
\usepackage{booktabs} 
\usepackage{array} 
\usepackage{makecell}
\usepackage{xcolor}
\usepackage{multirow}
\usepackage{mdframed}
\usepackage{tcolorbox}

\def\eg{e.g.} 

\def\ie{i.e.}
\def\etc{etc.}

\title{Learning Accurate Segmentation Purely from Self-Supervision}

\author{
    Zuyao You\textsuperscript{1},
    Zuxuan Wu\textsuperscript{1},
    Yu-Gang Jiang\textsuperscript{1}
}

\affiliation[1]{\mbox{Fudan University}}

\abstract{
\begin{abstract}

Accurately segmenting objects without any manual annotations remains one of the core challenges in computer vision. In this work, we introduce \textbf{Selfment}, a fully self-supervised framework that segments foreground objects directly from raw images without human labels, pretrained segmentation models, or any post-processing. Selfment first constructs patch-level affinity graphs from self-supervised features and applies NCut to obtain an initial coarse foreground-background separation. We then introduce \textbf{Iterative Patch Optimization (IPO)}, a feature-space refinement procedure that progressively enforces spatial coherence and semantic consistency through iterative patch clustering. The refined masks are subsequently used as supervisory signals to train a lightweight segmentation head with contrastive and region-consistency objectives, allowing the model to learn stable and transferable object representations. Despite its simplicity and complete absence of manual supervision, Selfment sets new state-of-the-art (SoTA) results across multiple benchmarks. It achieves substantial improvements on F\textsubscript{max} over previous unsupervised saliency detection methods on ECSSD (\(+4.0\%\)), HKUIS (\(+4.6\%\)), and PASCAL-S (\(+5.7\%\)). Moreover, without any additional fine-tuning, Selfment demonstrates remarkable zero-shot generalization to camouflaged object detection tasks (\eg, \(.910\) \(S_m\) on CHAMELEON and \(.792\) \(\mathcal{F}_{\beta}^{\omega}\) on CAMO), outperforming all existing unsupervised approaches and even rivaling the SoTA fully supervised methods.
\end{abstract}
}

\checkdata[Code]{\url{https://github.com/geshang777/Selfment}}
\pgfplotsset{compat=1.18}

\begin{document}
\maketitle

\vspace{-1.5em}

\input{section/introduction}

\input{section/relatedwork}
\input{section/methods}

\input{section/experiments}
\input{section/limitation}
\input{section/conclusion}

\bibliographystyle{plainnat}
\bibliography{main}

\end{document}

%% file: section/introduction.tex
\section{Introduction}
\label{sec:intro}

\begin{figure}[t]
\centering
\includegraphics[width=0.9\columnwidth]{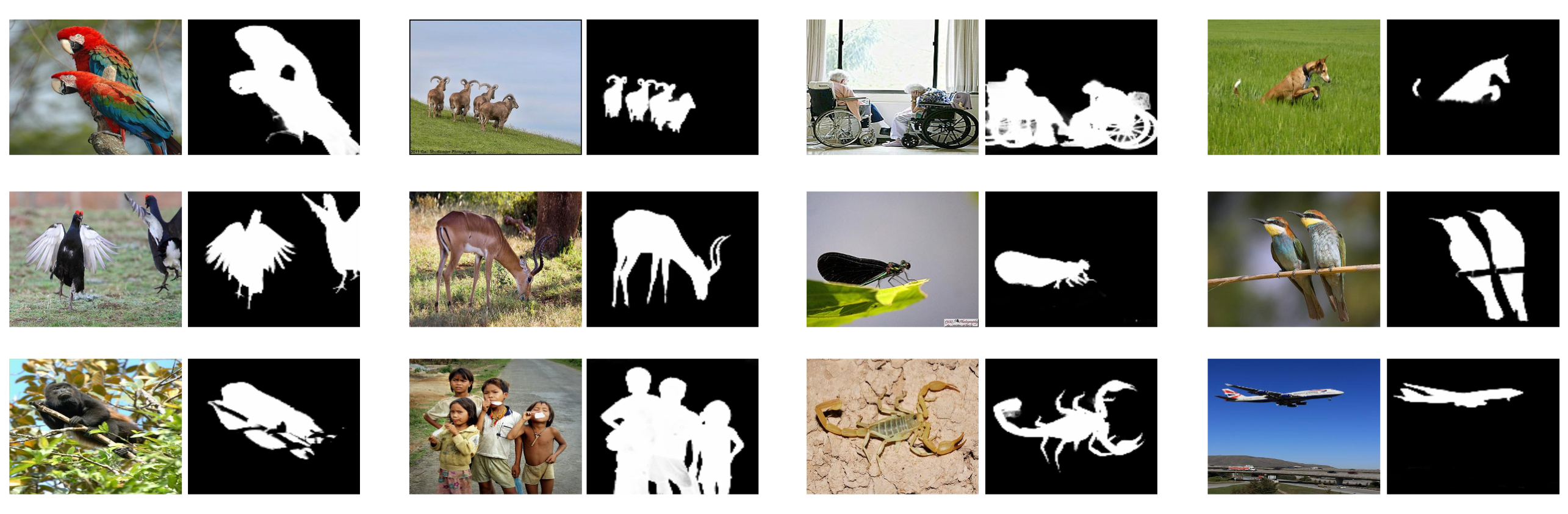}
\caption{We propose \textbf{Selfment}, a fully self-supervised framework for foreground segmentation that generates highly detailed and accurate saliency maps without any human-annotated labels or post-processing.}
\vspace{-1.3em}
\label{teaser}
\end{figure}

Object segmentation has long relied on dense, human-annotated masks~\cite{cheng2021per, cheng2021mask2former,lin2014microsoft}. While these annotations provide precise supervision, they are costly and time-consuming to collect, limiting scalability and relying heavily on the human inductive bias. To reduce annotation overhead, recent efforts~\cite{yuan2024unified, liu2024weakly, hu2024relax, hu2025int} have explored weakly supervised cues (\ie, points, scribbles, or motion trajectories, \etc) to guide segmentation. However, these methods still depend on human-provided signals and often rely on pretrained segmentation models (\eg, SAM~\cite{kirillov2023segment, ravi2024sam}) through fine-tuning or prompt adaptation. As a result, they remain partially tied to manual supervision and pretrained segmentation representations.

This raises a fundamental question:
Can a model learn accurate segmentation directly from unlabeled images without any human annotations or external off-the-shelf segmentation models?

Recent advances in self-supervised learning (SSL)~\cite{caron2021emerging,oquab2023dinov2,simeoni2025dinov3,he2022masked,bardes2024revisiting,zhou2021ibot} offer a promising path forward. SSL enables models to extract semantic structure directly from large-scale, unlabeled data. Among these methods, the DINO family~\cite{caron2021emerging,oquab2023dinov2,simeoni2025dinov3} is particularly notable for producing strong, object-centric representations. Trained via a teacher-student self-distillation framework, DINO models align features across diverse image crops, encouraging semantically consistent regions to share similar embeddings. DINOv3~\cite{simeoni2025dinov3} further advances this line of work by introducing Gram Anchoring to stabilize dense patch features over long training schedules, enabling the model to scale to 7B parameters while maintaining high-quality representations.

The dense feature maps produced by DINO-series naturally encode semantic similarity, such that patches belonging to the same object or category tend to have highly similar embeddings. This property provides strong semantic priors for downstream tasks such as segmentation or object discovery. Building on this idea, several recent works~\cite{simeoni2021localizing,wang2022self} treat the feature map as a graph and apply heuristic strategies, such as seed expansion or normalized graph cuts~\cite{shi2000normalizcssdd}, to partition it into foreground and background. While these methods can generate approximate object masks, the resulting bipartitions are unstable, and the resulting masks tend to be coarse. Achieving reasonable segmentation quality typically requires heavy post-processing (\eg, CRFs~\cite{krahenbuhl2011efficient}, bilateral solvers~\cite{barron2016fast}, or morphological refinements), which undermines the goal of the self-supervised segmentation.

In this work, we present \textbf{Selfment}, a fully self-supervised framework for foreground segmentation that requires no annotations, no post-processing, and no off-the-shelf segmentation models. Built on DINOv3~\cite{simeoni2025dinov3}, our approach begins by constructing a patch-wise affinity graph based on the image feature and applying Normalized Cut~\cite{shi2000normalizcssdd} to derive an initial bipartition, yielding a coarse yet semantically grounded segmentation prior. To further enhance spatial coherence and reduce noise from spectral relaxation, we introduce the \textbf{Iterative Patch Optimization (IPO)}, a simple yet effective module that refines assignments by iteratively clustering patches in the feature space. At each step, foreground and background centroids are updated, and patch labels are realigned according to semantic similarity. Orientation consistency constraints further stabilize the refinement, preventing degenerate solutions and ensuring that the evolving segmentation adheres to a coherent object-background separation. The refined masks then serve as self-supervised signals to train a lightweight segmentation head with \textbf{contrastive and region-consistency objectives}, enabling the model to progressively learn discriminative, object-aware representations.

Without any post-processing, Selfment surpasses the performance of previous state-of-the-art (SoTA) models by a very clear margin. On widely used salient object detection benchmarks such as ECSSD~\cite{shi2015hierarchical}, DUTS~\cite{wang2017learning}, HKUIS~\cite{li2015visual}, and PASCAL-S~\cite{li2014secrets}, Selfment yields substantial improvements of \(4.0\%\), \(7.0\%\), \(4.6\%\), and \(5.7\%\) in F\textsubscript{max}, respectively. As illustrated in Fig.~\ref{teaser}, Selfment can generate accurate and highly detailed saliency maps with the input resolution of \(2048\times2048\), purely based on our proposed self-supervised method. Furthermore, without any task-specific fine-tuning, Selfment demonstrates remarkable zero-shot generalization on camouflaged object detection tasks, achieving an \(S_m\) of \(.910\), \(.869\), \(.873\), and \(.902\) on CHAMELEON~\cite{skurowski2018animal}, CAMO~\cite{le2019anabranch}, COD10K~\cite{fan2020camouflaged}, and NC4K~\cite{lv2021simultaneously}, outperforming all previous unsupervised approaches and approaching the performance of fully supervised methods.

Our contributions are summarized as follows:
\begin{itemize}
\item We introduce \textbf{Selfment}, a fully self-supervised segmentation framework that operates without human annotations, external priors, or post-processing steps.
\item We introduce a simple yet effective mask refinement algorithm based on patch similarity, which significantly improves the performance of the initial NCut segmentation. Moreover, the method can be easily transferred across different self-supervised backbones.
\item Extensive experiments demonstrate that Selfment establishes new state-of-the-art results on both unsupervised salient object detection and camouflaged object detection tasks.

\end{itemize}

%% file: section/relatedwork.tex
\section{Related Work}
\label{sec:related_work}

\subsection{Self-supervised Vision Foundation Models}

Self-supervised vision foundation models~\cite{caron2021emerging,oquab2023dinov2,simeoni2025dinov3,he2022masked,lecun2022path} have advanced rapidly as backbone models trained without human annotations, enabling strong transfer to dense prediction tasks~\cite{you2025focus,cheng2021mask2former,wang2022self,simeoni2021localizing}. Early work~\cite{chen2020simple,he2020momentum,purushwalkam2020demystifying} explored contrastive objectives, which established that invariances learned from augmentations can yield competitive semantic features. Masked image modeling introduced an alternative paradigm~\cite{he2022masked,bao2021beit}, demonstrating that reconstructing masked patches produces scalable and robust backbone models. More recent foundation models~\cite{caron2021emerging,oquab2023dinov2,simeoni2025dinov3} have focused on dense, spatially consistent representations, with the latest DINOv3~\cite{simeoni2025dinov3} further strengthening dense feature stability through long-schedule training and Gram anchoring. Our work builds on this line by leveraging dense representations from DINOv3 and developing a fully self-supervised framework that converts these features into reliable foreground segmentation without manual masks.

\subsection{Unsupervised Object Segmentation}
Unsupervised object segmentation aims to separate foreground objects from background without relying on any manual annotations. Early unsupervised object segmentation methods relied heavily on low-level cues such as color contrast, texture consistency, motion boundaries, or super-pixel grouping~\cite{yuan2024unified, liu2024weakly,zhu2014saliency,li2015weighted}, which limited their robustness in complex scenes. More recent approaches alleviate these limitations by leveraging powerful off-the-shelf segmentation models such as SAM~\cite{kirillov2023segment} to provide pseudo labels or guidance~\cite{hu2024relax, hu2025int}, yet this dependence introduces strong external priors and reduces the level of true unsupervision. TokenCut~\cite{wang2022self} takes a different direction by using self-supervised ViT features and applying a normalized-cut formulation~\cite{shi2000normalizcssdd} to patch similarities, showing that object regions can emerge directly from transformer attention. However, its bipartitions are often unstable, and achieving competitive segmentation quality typically requires heavy post-processing, including bilateral solvers~\cite{barron2016fast} or CRFs~\cite{krahenbuhl2011efficient}. In contrast, our approach builds segmentation directly from self-supervised dense features, requires no annotated masks, does not rely on any external segmentation model such as SAM~\cite{kirillov2023segment,ravi2024sam}, and achieves high-quality foreground segmentation without any post-processing.

%% file: section/methods.tex
\section{Method}
\label{method}

\begin{figure*}[t]
\centering
\includegraphics[width=\columnwidth]{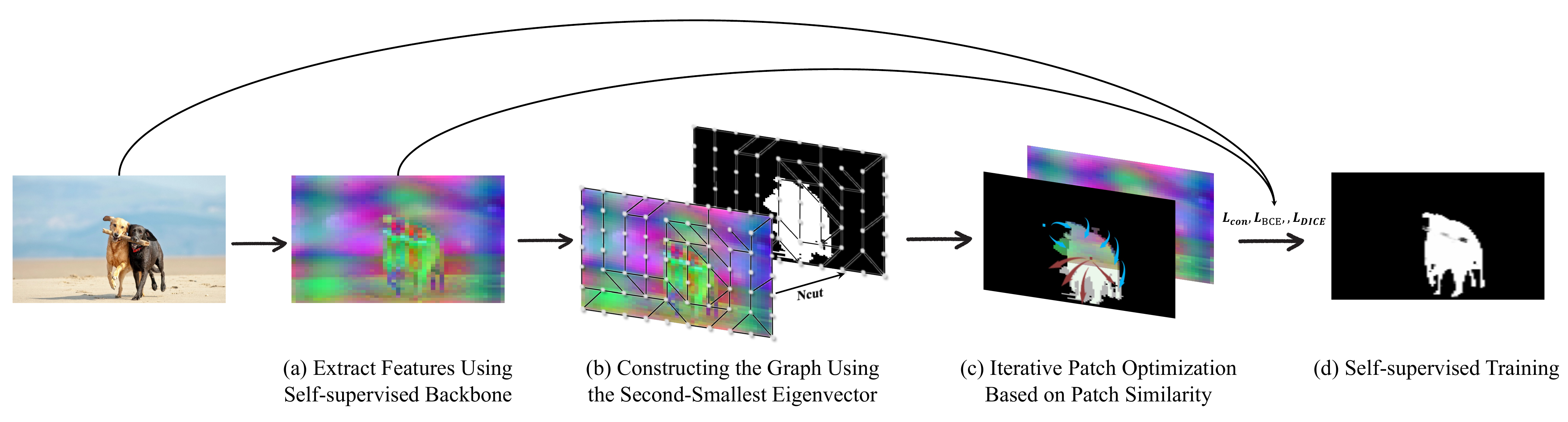}
\caption{An overview of Selfment. The input image is first encoded by a self-supervised backbone to produce dense patch features. These features define a patch-level affinity graph, from which we derive an initial foreground-background split using the second-smallest eigenvector of the NCut. We then apply Iterative Patch Optimization to improve spatial coherence and semantic consistency. The refined masks then serve as supervisory signals for training a lightweight segmentation head.}
\label{pipeline}
\end{figure*}

\subsection{Normalized Cut (NCut)}
\label{sec:ncut}

Given an image represented by a set of patch or pixel features 
$\{f_i\}_{i=1}^N$ extracted from the pretrained backbone, 
we construct an undirected weighted graph $G = (V, E)$, 
where each node $v_i \in V$ corresponds to a patch, 
and each edge $e_{ij} \in E$ represents the similarity between patches $i$ and $j$. 
The pairwise affinity is defined as
\begin{equation}
A_{ij} = 
\begin{cases}
\langle f_i, f_j \rangle, & \text{if } \langle f_i, f_j \rangle > \tau, \\
\epsilon, & \text{otherwise,}
\end{cases}
\label{eq:affinity}
\end{equation}
where $\tau$ is a similarity threshold set to \(0.2\) and $\epsilon$ is a small constant to ensure graph connectivity. 
Let $D$ denote the diagonal degree matrix with entries $D_{ii} = \sum_j A_{ij}$.

Following~\cite{shi2000normalizcssdd,wang2022self}, 
the NCut objective seeks to partition the graph into two disjoint sets 
$A$ and $B$ such that the inter-group similarity is minimized while maintaining 
high intra-group affinity:
\begin{equation}
\text{NCut}(A, B) = 
\frac{\text{cut}(A, B)}{\text{cut}(A, V)} +
\frac{\text{cut}(A, B)}{\text{cut}(B, V)},
\label{eq:ncut}
\end{equation}
where \(\text{cut}(X, Y) = \sum_{i \in X, j \in Y} X_{ij}\) measures the degree of similarity between two sets.

Minimizing $\text{NCut}(A,B)$ is equivalent to solving the generalized eigenvalue problem:
\begin{equation}
(D - A) \mathbf{x} = \lambda D \mathbf{x},
\label{eq:eigen}
\end{equation}
where $\mathbf{x}$ is a relaxed continuous indicator vector representing the partition. 
The smallest eigenvector corresponds to the trivial constant solution, while the 
second smallest eigenvector $\mathbf{x}_2$ (also known as the Fiedler vector) 
defines the optimal bipartition of the graph.

The binary segmentation mask is obtained by thresholding $\mathbf{x}_2$ at its mean value:
\begin{equation}
\text{mask}(i) = 
\begin{cases}
1, & \text{if } x_{2,i} > \frac{1}{N} \sum_{j=1}^{N} x_{2,j}, \\
0, & \text{otherwise.}
\end{cases}
\label{eq:mask}
\end{equation}

Finally, to identify the principal object, 
we locate the connected component that contains the seed patch 
(corresponding to the maximum absolute value in $\mathbf{x}_2$),
and treat it as the main object mask. 
This procedure yields a coarse foreground-background bipartition,
serving as the initialization for our subsequent refinement stage.

\subsection{Iterative Patch Optimization}
\label{sec:iterative_optimization}

The bipartition obtained from the normalized cut (Sec.~\ref{sec:ncut}) is often noisy and spatially inconsistent due to the binary graph construction and the spectral relaxation. 
To address this, we design an iterative refinement procedure that exploits the semantic similarity of patch features derived from self-supervised representations to progressively improve patch-level consistency.

Given the feature map $F = \{f_i \in \mathbb{R}^d\}_{i=1}^N$, 
we first normalize all patch embeddings by:
\begin{equation}
\tilde{f}_i = \frac{f_i}{\|f_i\|_2}.
\end{equation}
Let $\mathbf{y}^{(0)} \in \{0, 1\}^N$ denote the initial bipartition mask obtained from the NCut result, where 1 indicates foreground.
We compute initial cluster centroids for the foreground and background regions as
\begin{equation}
\mu_f^{(0)} = \frac{1}{|\mathcal{F}^{(0)}|} \sum_{i \in \mathcal{F}^{(0)}} \tilde{f}_i,
\quad
\mu_b^{(0)} = \frac{1}{|\mathcal{B}^{(0)}|} \sum_{i \in \mathcal{B}^{(0)}} \tilde{f}_i,
\end{equation}
where $\mathcal{F}^{(0)} = \{i \mid y^{(0)}_i = 1\}$ and $\mathcal{B}^{(0)} = \{i \mid y^{(0)}_i = 0\}$.

At each iteration $t$, we update the label of every patch according to its relative similarity to the current cluster means:
\begin{equation}
y^{(t+1)}_i = 
\begin{cases}
1, & \text{if } \langle \tilde{f}_i, \mu_f^{(t)} \rangle >
             \langle \tilde{f}_i, \mu_b^{(t)} \rangle, \\
0, & \text{otherwise.}
\end{cases}
\label{eq:iter_update}
\end{equation}
After relabeling, the new cluster means are recomputed as:
\begin{equation}
\begin{aligned}
\mu_f^{(t+1)} &= \frac{1}{|\mathcal{F}^{(t+1)}|}
                 \sum_{i \in \mathcal{F}^{(t+1)}} \tilde{f}_i,\\
\mu_b^{(t+1)} &= \frac{1}{|\mathcal{B}^{(t+1)}|}
                 \sum_{i \in \mathcal{B}^{(t+1)}} \tilde{f}_i.
\end{aligned}
\end{equation}

The process is repeated for a fixed number of iterations $T=20$.  
To avoid label flipping between iterations, we enforce orientation consistency by maintaining a reference vector 
$\mathbf{r} = \mu_f^{(0)} - \mu_b^{(0)}$, 
and reversing labels when 
$\langle (\mu_f^{(t+1)} - \mu_b^{(t+1)}), \mathbf{r} \rangle < 0$.

This iterative optimization refines the NCut initialization by aligning patch assignments with semantically similar features learned from self-supervised training. Our refinement relies solely on feature similarity, producing significantly cleaner and semantically coherent masks without any external priors or annotations.

\subsection{Self-supervised Training}
\label{sec:selfsupervised_training}

As illustrated in Fig.~\ref{pipeline}, we employ masks from Sec.~\ref{sec:iterative_optimization} as self-supervised signals to train a lightweight segmentation head that learns discriminative patch embeddings for more robust and stable unsupervised salient object detection.

We introduce a two-layer projection head followed by a binary classifier, denoted as $\phi_\theta$, that operates on patch features $f_i \in \mathbb{R}^d$ extracted from the vision backbone. The projection head maps features into an embedding space via:
\begin{equation}
z_i = \mathrm{\phi_\theta}(f_i) = W_2 \sigma(W_1 f_i + b_1) + b_2, 
\end{equation}
where $\sigma(\cdot)$ is a ReLU activation. The classifier then outputs logits 
$l_i = W_c z_i + b_c \in \mathbb{R}^2$ 
corresponding to foreground and background probabilities. 

For each image, we obtain patch-level pseudo-labels $\mathbf{y} \in \{0,1\}^N$ from the iterative patch optimization described in Sec.~\ref{sec:iterative_optimization}. These pseudo-labels provide noisy but spatially consistent foreground supervision that guides the patch embedding learning.

Inspired by InfoNCE~\cite{rusak2024infonce}, we impose a feature-level alignment encouraging embeddings of patches from the same region (foreground or background) to be close while pushing apart those from opposite regions. Let $z_i$ be $\ell_2$-normalized patch embeddings, and $S = z_i^\top z_j / \tau$ be their pairwise similarity with temperature $\tau$. We define:
\begin{equation}
\mathcal{L}_{\text{con}} = - \frac{1}{K} \sum_{i=1}^K 
\frac{1}{|\mathcal{P}_i|} \sum_{j \in \mathcal{P}_i} 
\log \frac{\exp(S_{ij})}
{\sum_{k \neq i} \exp(S_{ik})},
\end{equation}
where $\mathcal{P}_i = \{ j \neq i \mid y_j = y_i \}$ denotes positive pairs sharing the same label.
To further encourage segmentation consistency, we use the soft Dice loss:
\begin{equation}
\mathcal{L}_{\text{Dice}} = 1 - 
\frac{2 \sum_i p_i y_i + \epsilon}
{\sum_i p_i^2 + \sum_i y_i^2 + \epsilon},
\end{equation}
where $p_i = \sigma(l_i^{(1)})$ are the predicted foreground probabilities. Besides, each patch is trained to predict its pseudo-label via BCE loss:
\begin{equation}
\mathcal{L}_{\text{BCE}} = -\frac{1}{N} \sum_{i=1}^N 
\Big[ y_i \log \sigma(l_i^{(1)}) + (1-y_i) \log (1 - \sigma(l_i^{(1)})) \Big],
\end{equation}
where $\sigma(\cdot)$ denotes the sigmoid and $l_i^{(1)}$ is the foreground logit. The overall self-supervised loss is a weighted combination of these components:
\begin{equation}
\mathcal{L}_{\text{total}} = 
\lambda_{\text{con}} \mathcal{L}_{\text{con}} + 
\lambda_{\text{Dice}} \mathcal{L}_{\text{Dice}} +
\lambda_{\text{BCE}} \mathcal{L}_{\text{BCE}},
\end{equation}
where $\lambda_{\text{con}}$, $\lambda_{\text{Dice}}$, and $\lambda_{\text{BCE}}$ are set to \(0.1\), \(1.0\), and \(1.0\), respectively. 
This objective encourages the model to learn discriminative, spatially coherent patch embeddings that capture objectness purely from self-supervised cues, without any labeled data or downstream fine-tuning.

%% file: section/experiments.tex
\section{Experiment}
\label{experiment}

\begin{table*}[t!]
\centering
\caption{\textbf{Comparison of unsupervised saliency detection methods.} 
We evaluate Selfment against state-of-the-art unsupervised methods on ECSSD~\cite{shi2015hierarchical}, DUTS~\cite{wang2017learning}, HKUIS~\cite{li2015visual}, and PASCAL-S~\cite{li2014secrets}. 
Best results are \textbf{bolded}.}
\vspace{0.5em}
\resizebox{\textwidth}{!}{
\begin{tabular}{l|ccc|ccc|ccc|ccc}
\toprule
{} & \multicolumn{3}{c|}{ECSSD~\cite{shi2015hierarchical}} & \multicolumn{3}{c|}{DUTS~\cite{wang2017learning}} & \multicolumn{3}{c|}{HKUIS~\cite{li2015visual}} & \multicolumn{3}{c}{PASCAL-S~\cite{li2014secrets}} \\
\cmidrule(lr){2-4} \cmidrule(lr){5-7} \cmidrule(lr){8-10} \cmidrule(lr){11-13}
{} & $F_{\max}$ & IoU & Acc. & $F_{\max}$ & IoU & Acc. & $F_{\max}$ & IoU & Acc. & $F_{\max}$ & IoU & Acc. \\
\midrule
HS~\cite{yan2013hierarchical} & 67.3 & 50.8 & 84.7 & 50.4 & 36.9 & 82.6 & - & - & - & - & - & - \\
wCtr~\cite{zhu2014saliency} & 68.4 & 51.7 & 86.2 & 52.2 & 39.2 & 83.5 & - & - & - & - & - & - \\
WSC~\cite{li2015weighted} & 68.3 & 49.8 & 85.2 & 52.8 & 38.4 & 86.2 & - & - & - & - & - & - \\
DeepUSPS~\cite{nguyen2019deepusps} & 58.4 & 44.0 & 79.5 & 42.5 & 30.5 & 77.3 & - & - & - & - & - & - \\
BigBiGAN~\cite{voynov2021object} & 78.2 & 67.2 & 89.9 & 60.8 & 49.8 & 87.8 & - & - & - & - & - & - \\
E-BigBiGAN~\cite{voynov2021object} & 79.7 & 68.4 & 90.6 & 62.4 & 51.1 & 88.2 & - & - & - & - & - & - \\
LOST~\cite{simeoni2021localizing,shen2022learning} & 75.8 & 65.4 & 89.5 & 61.1 & 51.8 & 87.1 & - & - & - & - & - & - \\

TokenCut-\(768 \times768\)~\cite{wang2022self} & 87.8 & 75.9 & 92.9 & 73.0 & 60.5 & 90.3  & 82.2 & 64.0 & 92.1 &76.2 &61.3 &84.0\\
TokenCut-\(1280 \times1280\)~\cite{wang2022self} & 86.7 & 75.0 & 92.3 & 68.1 & 55.9 & 87.9  & 79.6 & 64.9 & 92.6 &80.0 &60.2 &83.2\\

SelfMask-\(768 \times768\)~\cite{shin2022selfmask} &91.9 &77.2 &93.9 &79.4 &62.4 &92.4 & 89.8 & 74.4 & 94.7 &86.0 &61.8 &85.5 \\
SelfMask-\(1280 \times1280\)~\cite{shin2022selfmask} & 89.9 & 72.1 & 92.1 & 68.1 & 55.9 & 87.9  & 88.2 & 69.6 & 93.4 &84.8 &58.5 &84.1\\

FOUND-\(768 \times768\)~\cite{simeoni2023found} & 91.4 & 78.9 & 94.4 & 76.4 & 64.8 & 93.7 & 87.7 & 68.4 & 93.8 & 85.4 & 63.7 & 86.3 \\
FOUND-\(1280 \times1280\)~\cite{simeoni2023found} &89.2 &71.9 & 92.5 &72.6 & 55.1 &91.5 &86.6 & 64.5 & 92.9 &84.0 &60.0 & 84.9  \\

\midrule
\textbf{Selfment (Ours)-\(768 \times768\)} & 95.3 & 82.4 & 95.2 & 85.1 & 66.6 & 91.8 & 93.9 & 80.0 & 95.8 & 91.5 & 71.2 & 88.8 \\
\textbf{Selfment (Ours)-\(1280 \times1280\)} & \textbf{95.9} & \textbf{84.3} & \textbf{95.8} & \textbf{86.4} &\textbf{ 68.4} & \textbf{92.3} & \textbf{94.4} & \textbf{81.6} & \textbf{96.1} & \textbf{91.7} & \textbf{71.6} & \textbf{88.8} \\

\bottomrule
\end{tabular}}
\label{tab:salient_detection}
\end{table*}

\begin{figure*}[h]
\centering
\includegraphics[width=\columnwidth]{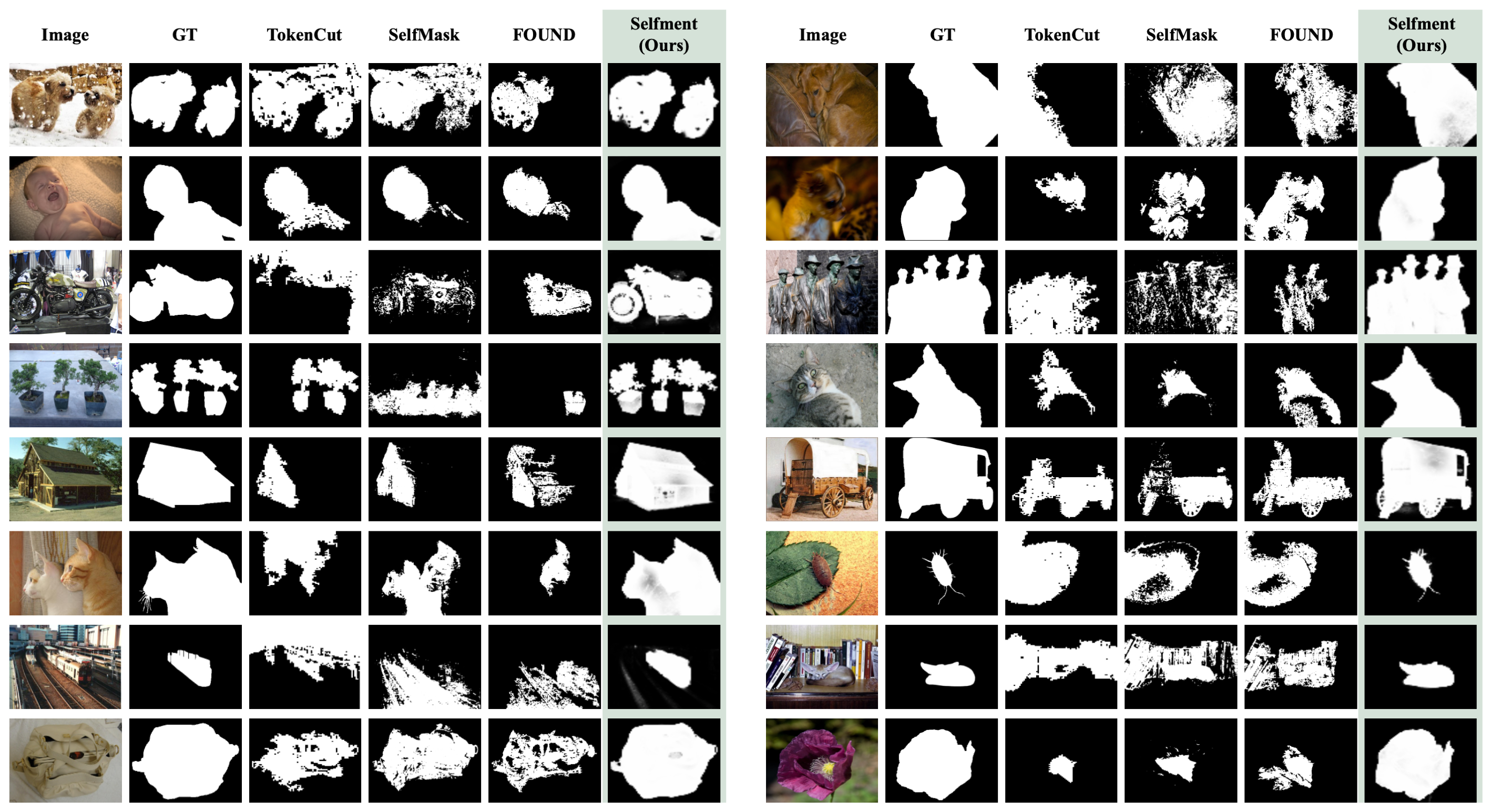}
\caption{Comparison with previous state-of-the-art methods on the unsupervised saliency detection task. All methods are evaluated  without any post-processing at an inference resolution of \(1280 \times 1280\). }
\label{comparison}
\end{figure*}

\begin{figure*}[h]
\centering
\includegraphics[width=0.95\columnwidth]{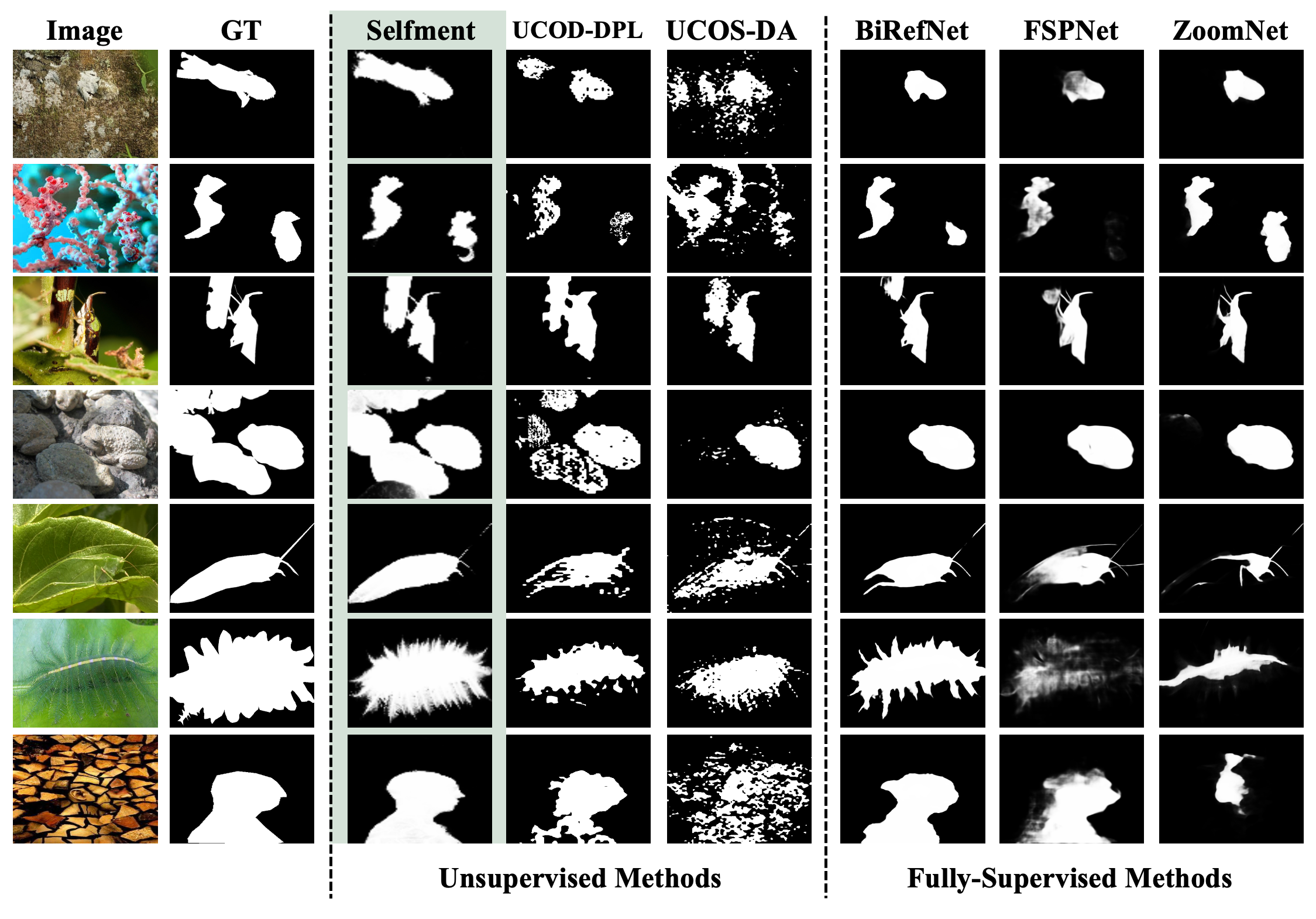}
\caption{Comparison with previous state-of-the-art on the camouflaged object detection tasks.}
\vspace{-7pt}

\label{comparison_cod}
\end{figure*}
\def\metricsCOD{

    &$\mathcal{S}_m\uparrow$
    &$\mathcal{F}_{\beta}^{\omega}\uparrow$
    & $E_{\xi}\uparrow$
    &$MAE$
}
\definecolor{myGray}{gray}{.92}
\begin{table*}[t!]
    \caption{\textbf{Comparison of camouflaged object detection methods.} 
We evaluate Selfment against state-of-the-art methods on CHAMELEON~\cite{skurowski2018animal}, CAMO~\cite{le2019anabranch}, COD10K~\cite{fan2020camouflaged}, and NC4K~\cite{lv2021simultaneously}. Best results are \textbf{bolded}.}
    \setlength{\belowcaptionskip}{0cm}   
    \renewcommand{\arraystretch}{1.0}
    \renewcommand{\tabcolsep}{2pt}
    \footnotesize
    \centering
\resizebox{\linewidth}{!}{
    \begin{tabular}{c|cccc|cccc|cccc|cccc}
        \toprule
        \multicolumn{1}{c}{\multirow{2}{*}[-1.2ex]{}} & \multicolumn{4}{c}{\textbf{CHAMELEON~\cite{skurowski2018animal}}} & \multicolumn{4}{c}{\textbf{CAMO~\cite{le2019anabranch}}}  & \multicolumn{4}{c}{\textbf{COD10K~\cite{fan2020camouflaged}}}& \multicolumn{4}{c}{\textbf{NC4K~\cite{lv2021simultaneously}}}\\
        \cmidrule[0.05em](lr){2-5} \cmidrule[0.05em](lr){6-9} \cmidrule[0.05em](lr){10-13} \cmidrule[0.05em](lr){14-17}
        \metricsCOD{}
        \metricsCOD{}
        \metricsCOD{}
        \metricsCOD{} \\
        \midrule
        \multicolumn{17}{c}{\textit{\textbf{Fully-Supervised Methods}}} \\ \midrule

        BGNet~\cite{sun2022boundary} &.901&.851&.954& .027 & .812 & .749 & .870  & .073 & .831 & .722 & .901 & .033 & .851 & .788   & .907 & .044 \\
        
        SINetv2~\cite{fan2021concealed} &.888&.816&.961&.030 & .820 & .743 & .882   & .070 & .815 & .680 & .887  & .037 & .847 & .770   & .903 & .048 \\
        ZoomNet~\cite{pang2022zoom}&.902&.845&.958&.023 & .820 & .752 & .878 & .066 & .838 & .729 & .888  & .029 & .853 & .784   & .896 & .043 \\
        FSPNet~\cite{huang2023feature} &.908&.851&.965&.023 & .856 & .799 & .899   & .050 & .851 & .735 & .895  & .026 & .879 & .816   & .915 & .035 \\
        BiRefNet~\cite{zheng2024bilateral} &.929&.911&.968&.016  & .932 & .914 & .974 & .015 & .913 & .874 & .960   & .014 & .914   & .894   & .953   & .023 \\ \midrule

        \multicolumn{17}{c}{\textit{\textbf{Semi-Supervised Methods}}} \\ \midrule
        CamoTeacher~\cite{lai2024camoteacher} & .756 & .617 & .813  & .065 & .701 & .560 & .795 & .112 & .759 & .594 & .854  & .049 & .791 & .687 & .868  & .068\\
        SCOD-ND~\cite{fu2024semi} & .850 & .773 & .928  & .036 & .789 & .732 & .859  & .077 & .819 & .725 & .891  & .033 & .838 & .787 & .903  & .046\\
        \midrule

        \multicolumn{17}{c}{\textit{\textbf{Unsupervised Methods}}} \\ \midrule
        BigBiGAN~\cite{voynov2021object} & .547 & .244 & .527  & .257 & .565 & .299 & .528  & .282 & .528 & .185 & .497  & .261 & .608 & .319 & .565  & .246 \\
        TokenCut~\cite{wang2022self} & .654 & .496 & .740  & .132 & .633 & .498 & .706  & .163 & .658 & .469 & .735  & .103 & .725 & .615 & .802  & .101\\
        SelfMask~\cite{shin2022selfmask} & .619 & .436 & .675  & .176 & .617 & .483 & .698  & .176 & .637 & .431 & .679  & .131 & .716 & .593 & .777  & .114\\

        UCOS-DA~\cite{zhang2023unsupervised} & .750&.639&.808&.091&.702&.604&.751&.148&.655&.467&.687&.120&.731&.617&.785&.103\\
        UCOD-DPL~\cite{yan2025ucod} & .864 & .825 & .931  & .031 & .793 & .747 & .862 & .077 & .834 & \textbf{.763} & \textbf{.916}  & \textbf{.031} & .850 & .818 & .923 & .043 \\
        \textbf{Selfment (Ours)} &\textbf{.910} &\textbf{.843} &\textbf{.944} &\textbf{.025} &\textbf{.869} &\textbf{.792} &\textbf{.894} &\textbf{.060} &\textbf{.873} &.754 &.900 & .033 &\textbf{.902} &\textbf{.836} &\textbf{.931} &\textbf{.033} \\
        \bottomrule
    \end{tabular}
 }
 
\label{tab:cod}
\end{table*}

\subsection{Implementation Details}
\label{sec:impl_details}

All experiments are implemented with DINOv3-7B and kept frozen during training. We randomly sample \(1{,}000\) images from the DUTS~\cite{wang2017learning} training set and use these as our self-supervised training corpus. The images are resized to the resolution of \(768 \times 768\) for the training stages. The patch-head (Sec.~\ref{sec:selfsupervised_training}) is optimized with Adam at a learning rate of $1\times10^{-3}$ for \(3\) epochs. To accelerate training, we extract and cache backbone patch features before head training and load these cached features from fast storage during each epoch. All experiments are carried out using distributed training on \(8\) NVIDIA A100 GPUs with \(80\)G memory with PyTorch DistributedDataParallel~\cite{paszke2019pytorch}. For reproducibility, we report averaged results where applicable.

\subsection{Benchmarks and Evaluation Metrics}

\paragraph{Unsupervised saliency detection.} Unsupervised saliency detection aims to identify and segment the most salient object in an image without supervision. We compare our method with a wide range of unsupervised saliency detection approaches.
Table~\ref{tab:salient_detection} reports the quantitative results on ECSSD~\cite{shi2015hierarchical}, DUTS~\cite{wang2017learning}, HKUIS~\cite{li2015visual}, and PASCAL-S~\cite{li2014secrets}. All the methods are compared in terms of \(F_{max}\), IoU, and pixel accuracy (Acc.) following previous work~\cite{wang2022self,shin2022selfmask,simeoni2023found}.

\paragraph{Camouflaged object detection (COD).} Camouflaged object detection is an extremely challenging segmentation task that requires identifying objects seamlessly concealed within their surroundings. We evaluate Selfment on CAMO~\cite{le2019anabranch}, COD10K~\cite{fan2020camouflaged}, NC4K~\cite{lv2021simultaneously}, and CHAMELEON~\cite{skurowski2018animal}, which are commonly use in COD tasks. We evaluate segmentation accuracy using S-Measure ($\mathcal{S}_m$), E-Measure ($E_{\xi}$), Weighted F-Measure ($\mathcal{F}_{\beta}^{\omega}$), and MAE following previous work ~\cite{pang2022zoom,zheng2024bilateral, fan2021concealed}.

\subsection{Main Results}
\label{sec:main_results}

\paragraph{Results on unsupervised saliency detection.}  
The works most similar to ours are~\cite{simeoni2021localizing, wang2022self, shin2022selfmask, simeoni2023found}, which also rely on feature embeddings from self-supervised foundations. For fair comparison, we standardize the input resolution during inference and compare these methods with Selfment at the resolution of \(768 \times768\) and \(1280 \times1280\). As shown in the Table.~\ref{tab:salient_detection}, Selfment outperforms these approaches across all metrics, achieving substantial improvements. Additionally, Selfment consistently benefits from increasing input image resolution during the inference stage. In contrast, other models tend to experience a decline in performance as the resolution increases. We further discuss this behavior in Sec.~\ref{sec:resolution}.

We also present a qualitative comparison between Selfment and previous methods in Fig.~\ref{comparison}. Benefiting from the iterative patch optimization and robust self-supervised training, Selfment produces saliency maps that are both complete and precise. In contrast, existing unsupervised methods tend to generate fragmented and incomplete predictions.

\paragraph{\textcolor{orange}{\emph{\textbf{Zero-shot}}} on COD.} We evaluate the zero-shot performance of Selfment on COD. Remarkably, Selfment not only surpasses all previous unsupervised methods by a large margin, but also outperforms several fully supervised methods. For example, as shown in the Table.~\ref{tab:cod}, Selfment achieves \textbf{\(.869\)} on $\mathcal{S}_m$ on CAMO~\cite{le2019anabranch}, significantly outperforming the previous unsupervised method (\(+ .076\)), and even surpassing strong fully supervised approaches such as FSPNet~\cite{huang2023feature}. Fig.~\ref{comparison_cod} further provides qualitative comparisons. Without any task-specific fine-tuning or post-processing, Selfment accurately detects camouflaged objects and produces detailed, high-quality saliency maps. These results demonstrate the strong generalization ability of Selfment and highlight the potential of self-supervised learning for challenging segmentation tasks such as COD.

\begin{figure*}[t]
\centering
\includegraphics[width=\columnwidth]{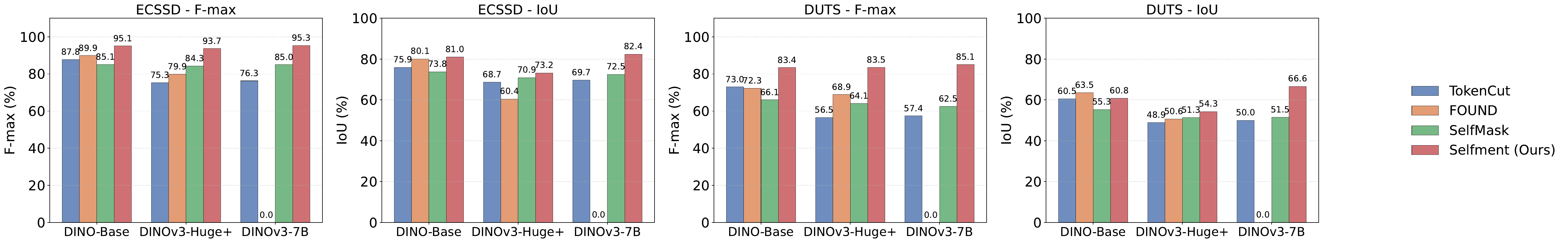}
\caption{Comparison of segmentation performance among TokenCut~\cite{wang2022self}, SelfMask~\cite{shin2022selfmask}, FOUND~\cite{simeoni2023found}, and Selfment using DINO-Base, DINOv3-Huge+, and DINOv3-7B as backbones. Metrics are reported on the ECSSD~\cite{shi2015hierarchical} dataset.
}
\vspace{-1.3em}

\label{fig:backbone_ablation}
\end{figure*}

\subsection{Ablation Study}
\label{sec:ablation}
\begin{wrapfigure}{r}{0.6\columnwidth}
    \centering
    \vspace{-15pt}
    \includegraphics[width=0.6\columnwidth]{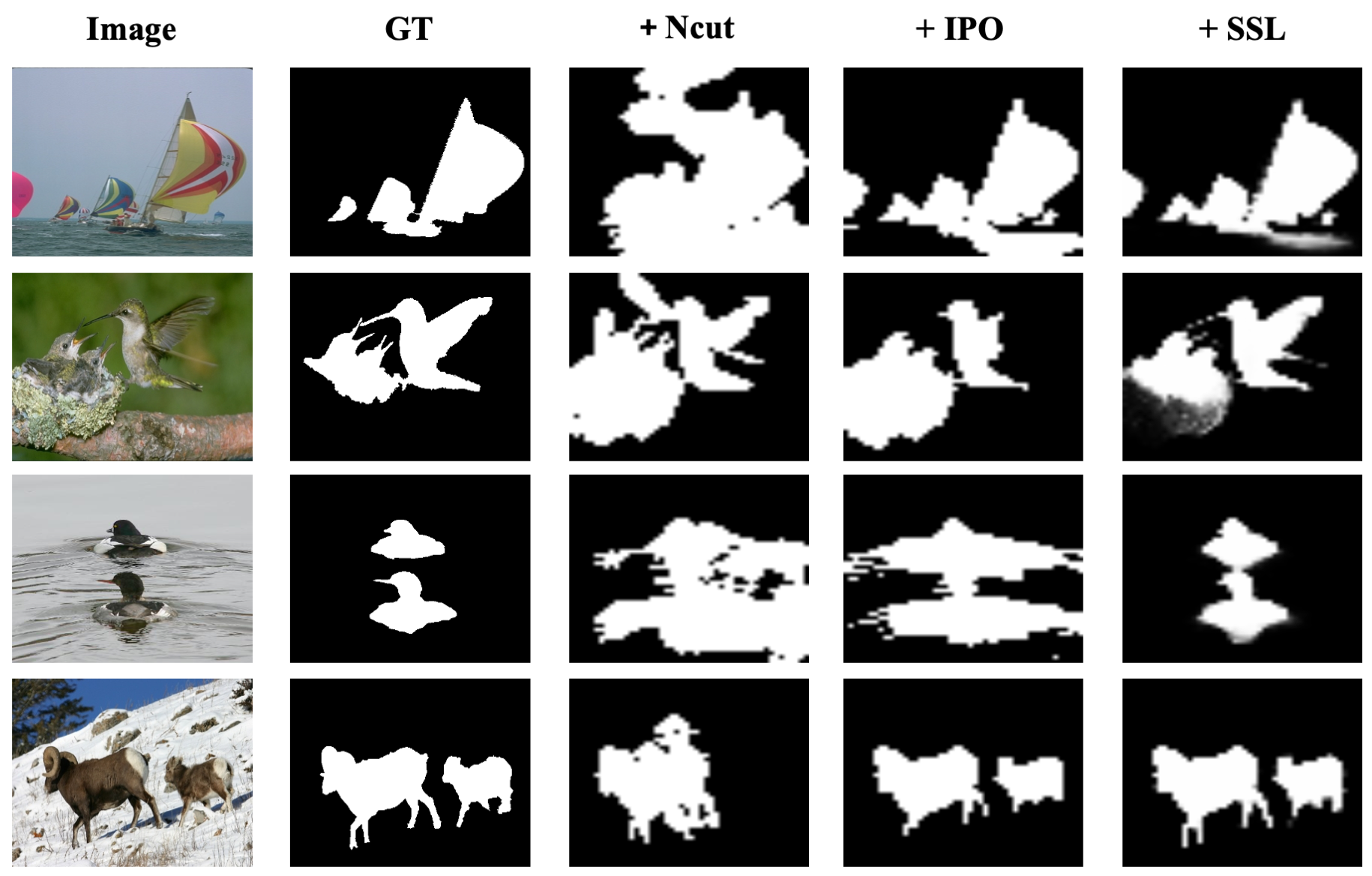}
    \caption{Ablation study on the main components of the Selfment pipeline. From left to right, we progressively add NCut, Iterative Patch Optimization (IPO), and self-supervised learning (SSL).}
    \label{method_ablation}
\end{wrapfigure}
We conduct ablation studies on the ECSSD~\cite{shi2015hierarchical} dataset to investigate the effect of our proposed modules and methods.
For the experiments reported in Table~\ref{tab:ablation}, we use a subset of \(500\) images sampled from DUTS for self-supervised training to allow agile iteration, fix the data sampling seed while keeping all other settings identical to those described in Sec.~\ref{sec:impl_details}.
The remaining experiments follow the same recipe as reported in Table~\ref{tab:salient_detection}. All experiments are conducted with strict variable control.

\paragraph{Ablation on Backbone}
\label{sec:backbone_ablation}

In this section, we extensively compare the performance of Tokencut~\cite{wang2022self}, SelfMask~\cite{shin2022selfmask}, FOUND~\cite{simeoni2023found}, and Selfment under the same backbone and resolution to explore the impact of different self-supervised foundation models on model performance. We use DINO-Base, DINOv3-Huge+, and DINOv3-7B~\cite{simeoni2025dinov3} as backbones. Notably, for SelfMask, due to the large size of the model, it cannot be trained on a single A100 GPU, so we used DeepSpeed Zero3~\cite{rasley2020deepspeed} for model sharding, while keeping other configurations unchanged. In addition, because the feature maps produced by DINO-Huge+ do not exhibit reliable bipartition properties under NCut, both TokenCut and Selfment use the value features from the last self-attention layer when DINO-Huge+ is adopted as the backbone. All experiments are conducted with a fixed inference resolution of \(768 \times 768\).

As shown in the Fig.~\ref{fig:backbone_ablation}, Selfment consistently outperforms previous methods by a clear margin, regardless of the backbone used. It is worth noting that, except for Selfment, the other models do not benefit from model scaling. In particular, FOUND exhibits instability during training when using DINOv3-7B as the backbone, resulting in both IoU and F-max being zero. This is because FOUND heavily relies on the initial background seed for similarity computation in the initialization step. However, due to the semantics of DINOv3 are very fine-grained, the similarity computed solely from the background seed leads to very fragmented pseudo-labels, making it difficult for the model to learn accurate semantics from the pseudo-labels, ultimately causing the training to fail. In contrast, Selfment effectively leverages the semantic structure encoded in both DINO and DINOv3 through components such as iterative patch optimization, enabling it to robustly exploit patch-level similarity and semantic consistency across different self-supervised backbones. Consequently, Selfment can achieve strong and stable performance regardless of backbone choice.

\begin{table}[t]
\centering
\caption{Ablation study of individual components. 
Each module is added cumulatively to the previous configuration. 
BCE: binary cross-entropy loss; Dice: soft Dice loss; Con.: contrastive similarity loss.}
\resizebox{0.8\linewidth}{!}{
\begin{tabular}{lccc|ccc}
\toprule
\textbf{Configuration} & \textbf{BCE} & \textbf{Dice} & \textbf{Con.} & $\mathbf{F_{\text{max}}}$ & \textbf{IoU} & \textbf{Acc}  \\
\midrule
NCut (baseline) &  &  &  & 74.7 & 63.9 & 86.2 \\
+ IPO &  &  & & 79.5 & 73.2 & 87.8  \\
\midrule
Self-supervised training & \checkmark &  &  & 88.3& 80.4 & 94.4  \\
\hspace{1.5em}+ Dice loss & \checkmark & \checkmark & & 88.9 & 81.3 & 94.7  \\
\hspace{1.5em}+ Contrastive loss & \checkmark &  & \checkmark & 88.9 & 81.4 & 94.7  \\
\hspace{1.5em}+ Dice \& Contrastive loss & \checkmark & \checkmark & \checkmark & \textbf{89.1} & \textbf{81.5} & \textbf{94.8}  \\
\bottomrule
\end{tabular}}
\label{tab:ablation}
\end{table}

\paragraph{Effect of IPO.}

\begin{wrapfigure}{r}{0.6\columnwidth}
    \centering
    \vspace{-15pt}
    \includegraphics[width=0.6\columnwidth]{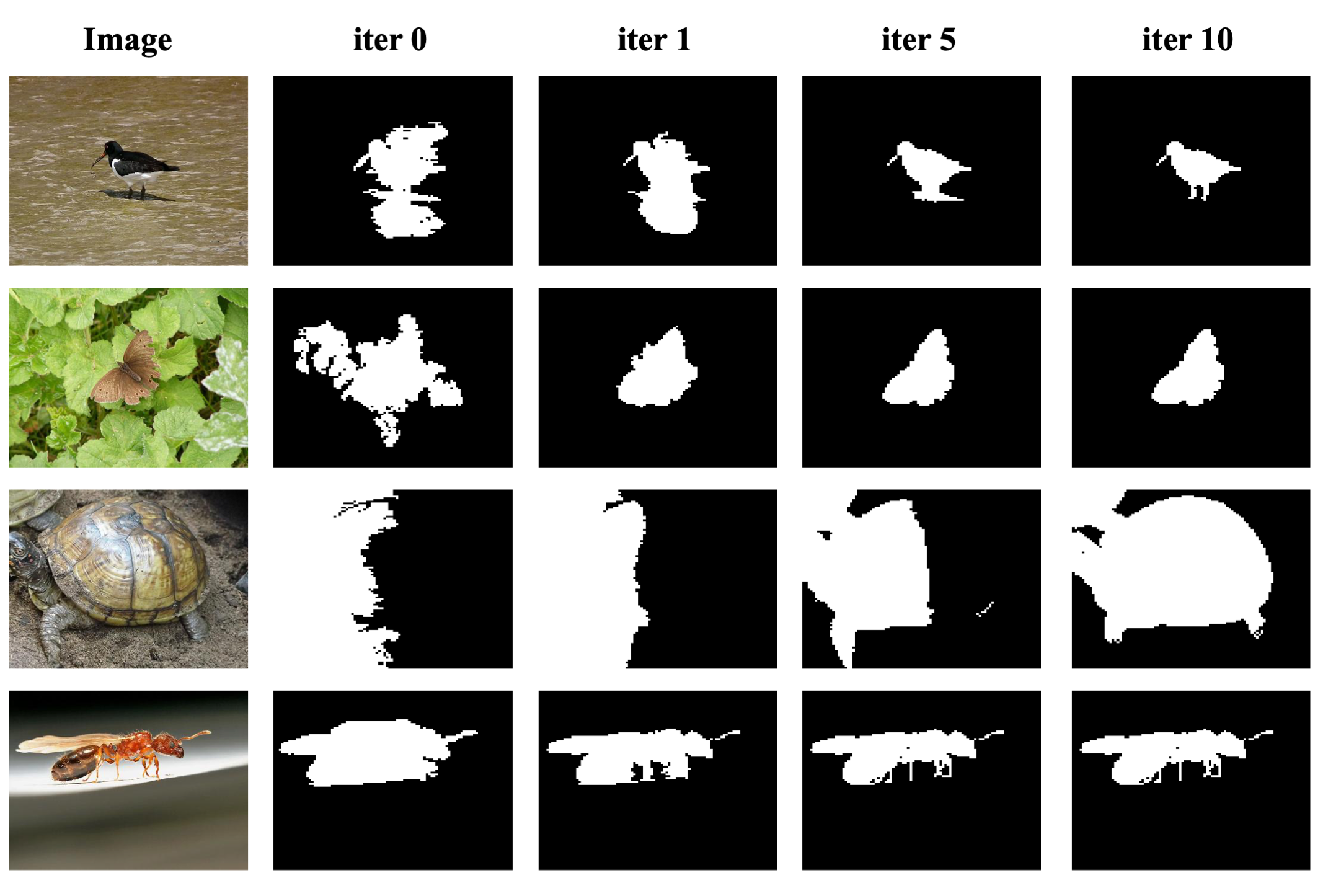}
    \caption{An illustration showing the progressive refinement of the mask across iterations of the IPO procedure.}
    \label{ipo_iter}
\end{wrapfigure}
IPO leverages feature similarity in the DINOv3 embedding space to iteratively refine patch assignments, producing more coherent and semantically meaningful object masks without any additional supervision.
As shown in Fig.~\ref{ipo_iter}, IPO substantially improves the ambiguous masks produced by the initial NCut step, quickly converging within roughly \(10\) iterations to a mask that closely aligns with the salient object boundary while preserving fine-grained details.
Quantitatively, introducing the iterative optimization stage yields a significant performance gain (Table~\ref{tab:ablation}), improving F\textsubscript{max}, IOU, and accuracy by \(4.8\%\), \(9.3\%\), and \(1.6\%\), respectively.

\paragraph{Self-supervised training.}  
We further employ the pseudo-labels generated by IPO to train a lightweight patch-level head in a self-supervised manner, enhancing the robustness and stability of the predicted saliency maps.
As shown in Table~\ref{tab:ablation}, training with these pseudo-labels significantly boosts segmentation accuracy.
Even when using only the BCE loss, the model reaches an F\textsubscript{max} of \(88.3\%\), a substantial improvement over the \(79.5\%\) obtained without the self-supervised training stage, indicating that the learned embeddings can effectively extract objectness cues from noisy pseudo-labels alone. Building on the BCE loss, we further incorporate the contrastive similarity loss and the Dice loss.
The contrastive similarity loss encourages the model to exploit patch-level feature relationships by pulling together patches belonging to the same region while pushing apart those from different regions.
Incorporating this loss leads to an additional performance gain, improving F\textsubscript{max} from \(88.3\%\) to \(88.9\%\) and IoU from \(80.4\%\) to \(81.4\%\).
In addition, adding the Dice loss promotes spatial compactness and boundary completeness, contributing a further \(+0.2\%\) improvement in IoU.

\paragraph{Comparison of Initial Bipartition Methods}
\label{sec:bipartition}

\begin{wraptable}{r}{0.45\columnwidth}
    \centering
    \vspace{-13pt}
    \caption{Comparison of initial bipartition methods on segmentation performance. Metrics reported on the ECSSD dataset.}
    \label{tab:initial_bipartition}

    \begin{tabular}{lccc}
        \toprule
        {} & $F_{\text{max}}$ & IoU & Acc \\
        \midrule
        \texttt{<CLS>} & 29.2 & 20.5 & 54.0 \\
        K-Means & 59.2 & 55.1 & 71.7 \\
        NCut & 74.7 & 63.9 & 86.2 \\
        \bottomrule
    \end{tabular}
\end{wraptable}
\noindent We compare different initial bipartition strategies in this section: using the DINOv3 \texttt{<CLS>} token, K-means, and NCut~\cite{shi2000normalizcssdd}, and report their performance on ECSSD~\cite{shi2015hierarchical} in terms of $\mathbf{F_{\text{max}}}$, IoU, and accuracy. As shown in Table.~\ref{tab:initial_bipartition}, using the \texttt{<CLS>} token to derive a binary partition performs poorly. The resulting bipartition yields only an $F_{\text{max}}$ of \(29.2\) on ECSSD. A simple K-means clustering over patch embeddings also offers a reasonable bipartition, but its quality remains significantly worse than that of NCut. In contrast, NCut provides a substantially more reliable initialization. Leveraging the second smallest eigenvector to guide the partition produces a meaningful foreground--background separation, achieving \(74.7\) $F_{\text{max}}$, \(63.9\) IoU, and \(86.2\) accuracy.

\paragraph{Effect of Input Resolution.}
\label{sec:resolution}

\begin{wrapfigure}{r}{0.7\columnwidth}
    \centering
    \vspace{-15pt}
    \includegraphics[width=0.7\columnwidth]{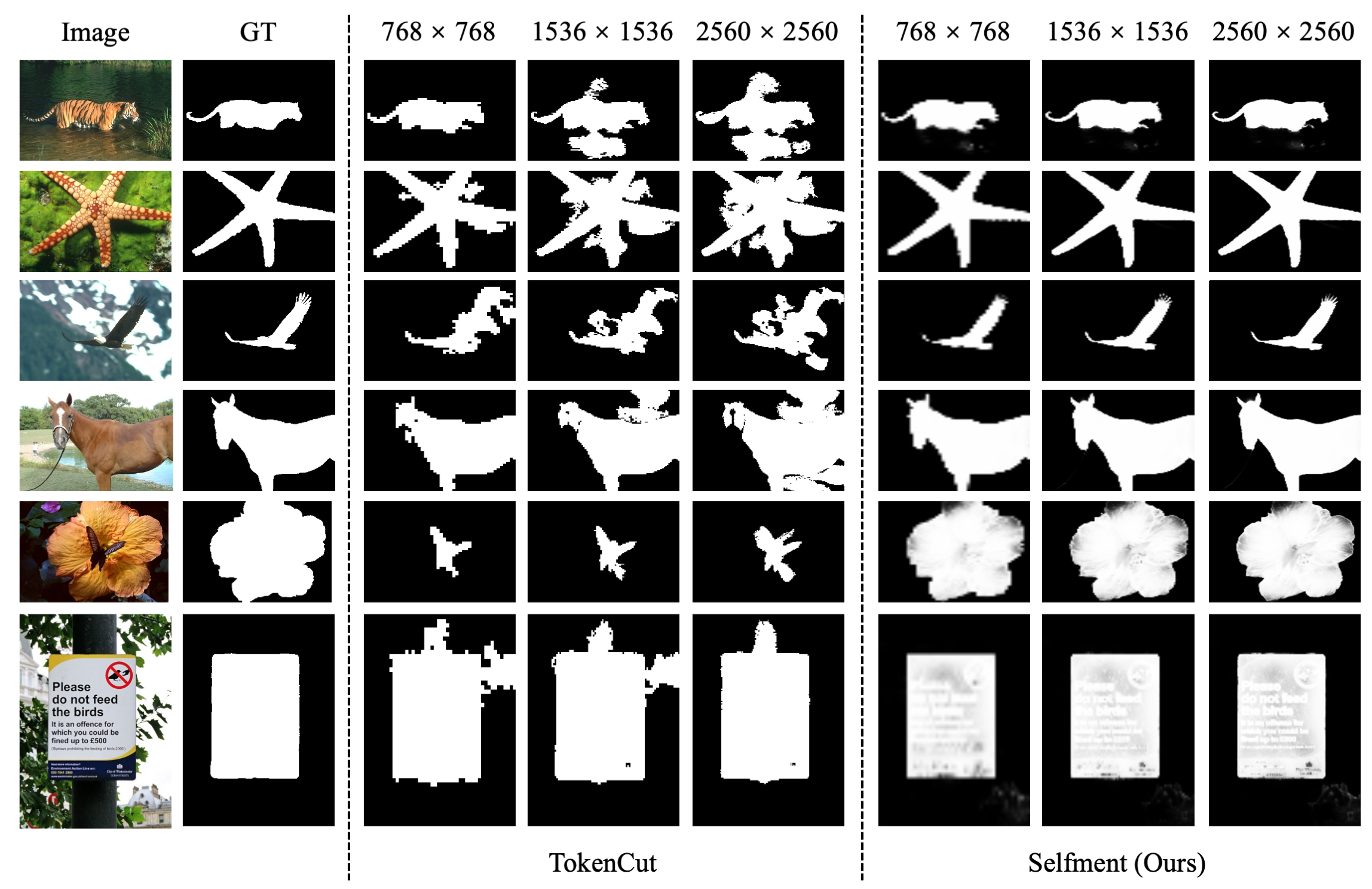}
    \caption{Qualitative comparison of how Selfment and TokenCut behave as input resolution increases.}
    \label{resulotion}
\end{wrapfigure}
In this section, we analyze how Selfment and TokenCut~\cite{wang2022self} behave as the input resolution increases in the Figure.~\ref{resulotion}. For a fair comparison, both methods operate on the output feature from the last layer of the DINOv3-7B. We evaluate saliency map quality by resizing the input images to three resolutions: \(768\times768\), \(1536\times1536\), and \(2560\times2560\). Although Selfment is trained only at \(768\times768\) resolution, it generalizes naturally to much higher resolutions. As the input size increases, the predicted saliency maps become progressively sharper and more detailed, while maintaining object-level coherence, \eg, the text on the signboard in the last row. TokenCut, however, depends heavily on a single NCut bipartition, which becomes unstable on high-resolution affinity graphs. As the resolution grows, this instability leads to noticeable degradation in its predictions.

\subsection{Computational Efficiency}
\label{sec:computational_efficiency}

During training, we resize all images to \(768 \times 768\). Unless otherwise specified, all experiments adopt the DINOv3-7B backbone. The backbone remains frozen throughout training, and only the lightweight segmentation head described in Sec.~\ref{sec:selfsupervised_training} is optimized. This head contains merely \(0.54\)M trainable parameters and requires \(1.08\)M FLOPs per forward pass. Furthermore, we cache the backbone features to avoid redundant computation. Using \(8\)$\times$A100 GPUs, Selfment is trained for \(3\) epochs in only \(27.6\) minutes.  
During inference, a single A100 GPU processes an \(768 \times 768\) image in \(2.69\) s with cached features and \(6.62 \)s without caching, which remains practical given the scale of the DINOv3-7B.

%% file: section/limitation.tex
\section{Limitation}
\label{limitation}

\begin{wrapfigure}{r}{0.5\columnwidth}
    \vspace{-13pt}
    \centering
    \includegraphics[width=0.5\columnwidth]{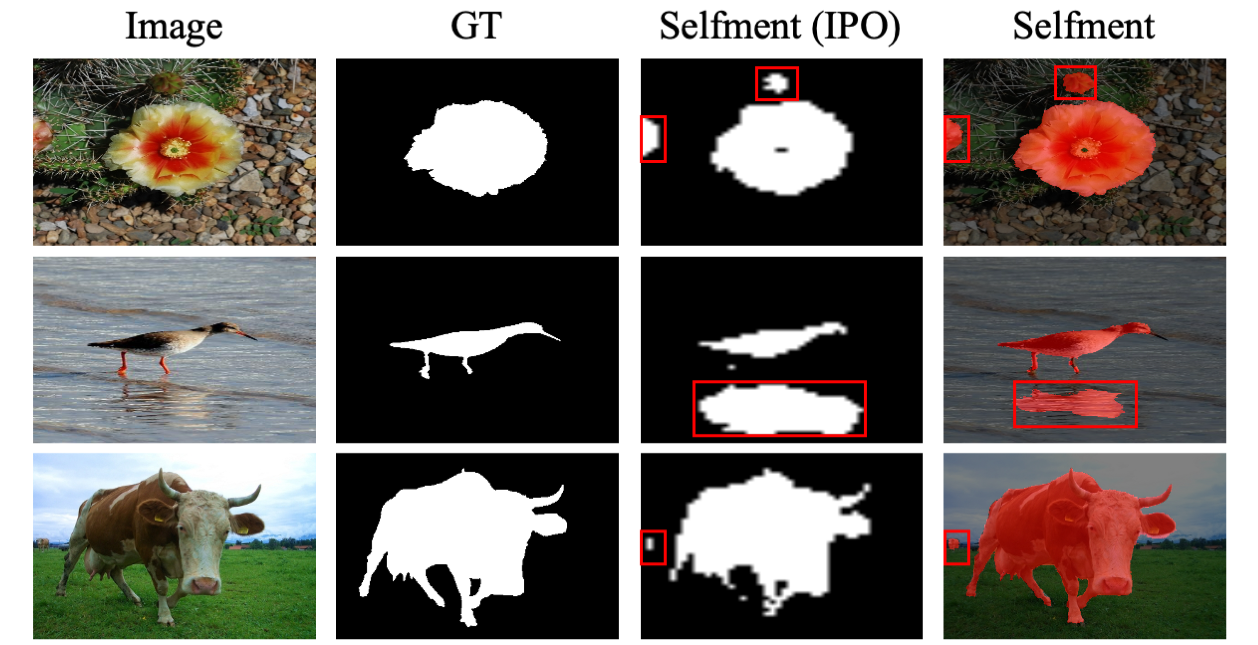}
    \caption{Failure cases on unsupervised saliency detection.}
    \vspace{-5pt}
    \label{badcase}
\end{wrapfigure}

Although Selfment achieves strong performance on unsupervised tasks, failure cases still exist, as illustrated in Figure~\ref{badcase}.
Because the IPO relies on patch-level feature similarity, the model sometimes incorrectly classifies objects that are semantically similar to the foreground as part of the foreground mask, as highlighted by the red boxes. Consequently, pseudo-masks generated for self-supervised training inherit the same errors. Developing more robust methods to leverage self-supervised backbone features for downstream tasks to fully leverage the rich semantic representations from DINOv3 represents a promising direction for future research.

%% file: section/conclusion.tex
\section{Conclusion}
\label{conclusion}

In this work, we presented Selfment, a fully self-supervised segmentation framework that achieves state-of-the-art performance without relying on human annotations, post-processing, or priors from SAM. By leveraging self-supervised feature learning and an iterative patch optimization method, Selfment produces highly accurate saliency maps and outperforms existing approaches on multiple benchmarks. Additionally, it can be easily transferred to camouflage object detection tasks, significantly surpassing previous methods. Selfment demonstrates that high-quality segmentation can be achieved entirely through self-supervision, setting a new standard for fully autonomous, annotation-free segmentation.